\documentclass[conference]{IEEEtran}
\IEEEoverridecommandlockouts

\usepackage{amsmath,amssymb,amsthm}
\usepackage{booktabs}
\usepackage{graphicx}
\usepackage{balance}

\newtheorem{lemma}{Lemma}
\newtheorem{remark}{Remark}

\begin{document}

\title{Predicting Estimated Times of Restoration for Electrical Outages Using Longitudinal Tabular Transformers}

\author{%
\IEEEauthorblockN{Sai Prasanna Teja Bogireddy\IEEEauthorrefmark{1},
                  Valliappan Muthukaruppan\IEEEauthorrefmark{1},
                  and Benjamin Carls\IEEEauthorrefmark{1}}
\IEEEauthorblockA{\IEEEauthorrefmark{1}Exelon Corporation, Chicago, IL, USA\\
teja.bogireddy@exeloncorp.com,
valliappan.muthukaruppan@exeloncorp.com,
benjamin.carls@comed.com}%
}

\maketitle

\begin{abstract}
Utilities publish Estimated Times of Restoration (ETRs) for customer-facing
storm outages, and their accuracy governs whether customers can make sound
decisions about food, medical equipment, and relocation. Prior work treats
ETR as static tabular regression in which each outage contributes one record,
discarding the fact that every development of an outage, from crew assignment
through dispatch, suspension, damage assessment and partial restoration, is
recorded as a revision. We reformulate ETR prediction as longitudinal tabular
regression and introduce a Longitudinal Tabular Transformer (LTT), an
axial-attention model that consumes the revisions preceding a prediction and
issues a refined estimate at every one. On 242{,}928 storm-attributed outages
from a cohort of 526{,}468 filtered events and 10.0 million revisions at six
operating companies, LTT reduces customer-weighted asymmetric error at all
six, by a median of 36.9\,\% against the estimates the utilities published
during the same storms and 11.3\,\% against the strongest learned baseline at
each. It is the only method improving on the incumbent's satisfaction impact
at all six companies while also reducing root mean squared error at all six.
Stratification by revision index shows that LTT error is largest at the first
revision, where no history is available, and falls monotonically as revisions
accumulate.
\end{abstract}


\section{Introduction}
The ETR is the primary information a utility conveys to an affected customer:
households with powered medical equipment, businesses weighing refrigerated
inventory, and municipalities deciding whether to open warming centres all
act on it. Utilities generate ETRs by manual dispatcher assessment or
rule-based OMS heuristics \cite{adibi,wanik}, and research has framed the
task as tabular regression over a feature vector \cite{walsh,kar}. Both treat
an outage as a static object, characterised once and predicted once.

An outage is not static. A crew is assigned, then dispatched; it arrives and
reports the damage; it is suspended and redirected to a higher-priority job;
some customers are restored while the rest await a second repair. Each
development is written to the Outage Management System (OMS) as a revision
and each changes the expected remaining duration, so a model observing only
the latest revision discards the trajectory. We instead predict remaining
time at every revision, conditioned on those preceding it.

\textbf{Scope of novelty.} Weather-driven outage modelling has concentrated
on outage \emph{counts} \cite{nateghi}; restoration duration has received
less attention \cite{adibi,wanik,walsh,kar} and that work is static.
Attention over feature tokens within a tabular record \cite{fttransformer},
the axial factorisation of feature and time axes \cite{axial}, and
transformers over tabular sequences in other domains \cite{tabbert} are all
established; we claim none of them. We claim that no prior work models the
revision stream of an outage and none benchmarks against the estimates a
utility published. We contribute the formulation and the LTT architecture; a benchmark against
customer-facing OMS estimates with an explicit cadence-matched control; and
replication at six operating companies, with model error localised by revision
index.

\section{Problem Formulation}
Let $\mathcal{E}=\{e_1,\dots,e_N\}$ be outage events, $e_i$ comprising
revisions $r_{i,1},\dots,r_{i,M_i}$ at times $t_{i,1}<\dots<t_{i,M_i}$, each
carrying categorical features $C_{i,j}$, continuous features $X_{i,j}$, and
system context $Z_{i,j}$. We predict remaining time to restoration
\begin{equation}
y_{i,j}=T^{\mathrm{restore}}_i-t_{i,j},
\label{eq:target}
\end{equation}
recovering the published estimate as
$\widehat{\mathrm{ETR}}_{i,j}=t_{i,j}+\hat y_{i,j}$. Predicting remaining time
rather than an absolute restoration timestamp removes a deterministic
component of the target: a model predicting $T^{\mathrm{restore}}_i$ must
reproduce the current clock and add a correction to it, so the scale of the
target is set by calendar position and its support shifts between storms. A
remaining-time target is invariant to when the outage occurred and directly
comparable across events, leaving only the uncertain quantity to be learned.
Prediction at revision $j$ conditions on the trailing window
\begin{equation}
\mathcal{W}(i,j)=\bigl(r_{i,j-W+1},\dots,r_{i,j}\bigr),
\label{eq:window}
\end{equation}
whose slot $a\in\{1,\dots,W\}$ holds $r_{i,j-W+a}$, with the convention that
$r_{i,k}$ denotes a null token for $k\le 0$, so \eqref{eq:window} is
well-defined for every $j\ge 1$. Writing $w=\min(j,W)$ for the number of
observed revisions in the window, the observation mask
\begin{equation}
m^{\mathrm{obs}}_a=\mathbb{I}[a>W-w],\qquad a=1,\dots,W,
\label{eq:obsmask}
\end{equation}
marks slot $a$ as observed. Padding therefore occupies the leading $W-w$
slots and the target revision always occupies $a=W$, which is what allows the
head of Section~III to read a fixed index. We set $W=20$. The window slides
with the target, so revision forty is predicted from revisions twenty-one
through forty and only the earliest history is discarded; $73.5\,\%$ of
outages have twenty revisions or fewer (range $58.8$--$96.1\,\%$).

\begin{table*}[t]
\centering
\caption{Cohort summary by operating company, ordered by median revisions per outage. Storm counts are fused district
windows; cohort events outside declared storm windows contribute context only.
Revision counts are cohort-wide.}
\label{tab:cohort}
\footnotesize
\setlength{\tabcolsep}{5pt}
\begin{tabular}{lcccrrrcc}
\toprule
& \multicolumn{3}{c}{Storms} & \multicolumn{2}{c}{Events}
& & \multicolumn{2}{c}{Revisions/event} \\
\cmidrule(lr){2-4}\cmidrule(lr){5-6}\cmidrule(lr){8-9}
OPCO & Train & Val & Test & Storm & Cohort & Revisions
& Median & p95 \\
\midrule
OPCO-1 & 50 & 5 & 9 & 109{,}541 & 190{,}053 & 5{,}328{,}117 & 18 & 67 \\
OPCO-2 & 67 & 5 & 9 &  61{,}940 & 101{,}797 & 1{,}953{,}174 & 12 & 62 \\
OPCO-3 & 13 & 3 & 3 &   8{,}140 &  38{,}560 &   434{,}711 & 10 & 26 \\
OPCO-4 & 35 & 5 & 9 &  10{,}421 &  53{,}481 &   689{,}865 & 10 & 30 \\
OPCO-5 & 50 & 5 & 9 &  23{,}387 &  90{,}845 & 1{,}112{,}829 & 10 & 29 \\
OPCO-6 & 81 & 5 & 9 &  29{,}499 &  51{,}732 &   490{,}813 &  8 & 19 \\
\midrule
Total & 296 & 28 & 48 & 242{,}928 & 526{,}468 & 10{,}009{,}509 & 12 & 51 \\
\bottomrule
\end{tabular}
\end{table*}

\section{Longitudinal Tabular Transformer}
\textbf{Tokenization.} Each feature is embedded independently into
$\mathbb{R}^d$ \cite{fttransformer}: for continuous feature $k$ with value
$x^{(k)}_{i,j}$, and for categorical feature $k$ at level $c^{(k)}_{i,j}$
from vocabulary $V_k$,
\begin{equation}
u^{(k)}_{i,j}=x^{(k)}_{i,j}w_k+b_k,
\qquad
u^{(k)}_{i,j}=E_k\bigl[c^{(k)}_{i,j}\bigr],
\label{eq:tok}
\end{equation}
$w_k,b_k\in\mathbb{R}^d$, $E_k\in\mathbb{R}^{|V_k|\times d}$. Missingness is
represented explicitly rather than imputed: where the indicator
$m^{(k)}_{i,j}$ is active a learned vector is added,
$u^{(k)}_{i,j}\!\leftarrow\!u^{(k)}_{i,j}+m^{(k)}_{i,j}\mu_k$. An absent
crew-assignment timestamp records that no crew has yet been assigned, so the
absence is itself informative. Stacking gives
$U_{i,j}\in\mathbb{R}^{F\times d}$, $F=p+q=54$.

\textbf{Intra-revision axis.} With $\mathrm{TB}(A)=\tilde A+\mathrm{FFN}
(\mathrm{LN}(\tilde A))$ and $\tilde A=A+\mathrm{MHA}(\mathrm{LN}(A))$, and a
learned pooling token $\rho$ prepended at index $0$ so that
$U^{(0)}_{i,j}=[\rho;U_{i,j}]$,
\begin{equation}
U^{(\ell+1)}_{i,j}=\mathrm{TB}^{\mathrm{feat}}\bigl(U^{(\ell)}_{i,j}\bigr),
\quad \ell=0,\dots,L_f-1,
\label{eq:feataxis}
\end{equation}
with $h_{i,j}=U^{(L_f)}_{i,j}[0]$. Parameters are shared across positions and
events; joint attention over the $W\times F$ grid needs $O(W^2F^2)$ pairwise
terms against $O(WF^2+W^2)$ factorised. Padded positions are zeroed and, by
\eqref{eq:mask}, never attended.

\textbf{Temporal encoding.} Revisions arrive at irregular intervals set by
operational activity, so positional index alone is inadequate. We apply a
learned periodic representation \cite{time2vec} to six channels
$\tau_{i,j}$:
\begin{equation}
\phi(\tau)_n=
\begin{cases}
\omega_n\tau+\varphi_n, & n=0,\\
\sin(\omega_n\tau+\varphi_n), & 1\le n<d_t,
\end{cases}
\label{eq:t2v}
\end{equation}
$\{\omega_n,\varphi_n\}$ learned; the linear term preserves monotone elapsed
time, the periodic terms capture diurnal crew scheduling. Concatenating and
projecting gives $\pi_{i,j}=W_\tau[\phi(\tau^{(1)}_{i,j});\dots;
\phi(\tau^{(6)}_{i,j})]$. Channels are elapsed time since the outage opened,
since the previous revision, hour of day, time since storm onset, a causal
proxy for storm peak, and normalised window position; the peak proxy uses the
running maximum of customers interrupted to $t_{i,j}$, the realised peak
being unobservable at prediction time.

\textbf{Static, context, and inter-revision axes.} Static features are
tokenized as in \eqref{eq:tok} and mean-pooled to $\sigma_i$, prepended
rather than broadcast, since replicating a constant across $W$ positions
consumes attention capacity without conveying information. Context is
projected per position into $G_{i,j}\in\mathbb{R}^{W\times d}$. The sequence
\begin{equation}
H^{(0)}_{i,j}=\bigl[\sigma_i;\,h_{i,j-W+1}+\pi_{i,j-W+1};\dots;
h_{i,j}+\pi_{i,j}\bigr]
\label{eq:seq}
\end{equation}
in $\mathbb{R}^{(W+1)\times d}$, indexed $0,\dots,W$, receives one
cross-attention layer against the context, $H^{(0)}\!\leftarrow\!H^{(0)}
+\mathrm{MHA}(\mathrm{LN}(H^{(0)}),G,G;M^{\mathrm{ctx}})$, followed by
\begin{equation}
H^{(\ell+1)}_{i,j}=\mathrm{TB}^{\mathrm{rev}}\bigl(H^{(\ell)}_{i,j},M\bigr),
\quad \ell=0,\dots,L_r-1.
\label{eq:revaxis}
\end{equation}
The always-present static token extends \eqref{eq:obsmask} by $\tilde m_0=1$,
$\tilde m_a=m^{\mathrm{obs}}_a$ for $1\le a\le W$, giving
\begin{equation}
M_{ab}=\begin{cases}0,&\tilde m_b=1\\ -\infty,&\text{else}\end{cases}
\qquad
M^{\mathrm{ctx}}_{ab}=\begin{cases}0,&m^{\mathrm{obs}}_b=1\\
-\infty,&\text{else,}\end{cases}
\label{eq:mask}
\end{equation}
$M\in\mathbb{R}^{(W+1)\times(W+1)}$, $M^{\mathrm{ctx}}\in
\mathbb{R}^{(W+1)\times W}$, added to the logits before the softmax. Both
depend only on the key index $b$ and are key-padding masks, not general
attention patterns.

\begin{remark}[Absence of temporal leakage]
No causal mask is required. Every position of $\mathcal{W}(i,j)$ indexes
$r_{i,j'}$ with $j'\le j$, hence $t_{i,j'}\le t_{i,j}$ and
$r_{i,j'}\in\Omega_{i,j}$, the information available at $t_{i,j}$; so
$\hat y_{i,j}$ is a measurable function of $\Omega_{i,j}$ irrespective of the
attention pattern applied within the window. Bidirectional attention is
therefore admissible and strictly more expressive. This does not hold for the
alternative in which one pass over an entire event is supervised at every
position, which does require causal masking.
\end{remark}

\textbf{Head.} With $v_{i,j}=\psi(H^{(L_r)}_{i,j}[W])$ we predict ordered
levels $Q=\{\kappa_1,\kappa_2,\kappa_3\}=\{\tfrac12,\tfrac56,\tfrac9{10}\}$
as a base plus non-negative increments,
\begin{align}
\hat y^{(\kappa_1)}_{i,j}&=s\cdot\mathrm{softplus}
   \bigl(a_1^\top v_{i,j}+b_1\bigr),\label{eq:head1}\\
\hat y^{(\kappa_m)}_{i,j}&=\hat y^{(\kappa_{m-1})}_{i,j}
   +s\cdot\mathrm{softplus}\bigl(a_m^\top v_{i,j}+b_m\bigr),\; m=2,3.
\label{eq:head2}
\end{align}
Non-negativity follows from the softplus and monotonicity
$\hat y^{(\kappa_1)}\!\le\!\hat y^{(\kappa_2)}\!\le\!\hat y^{(\kappa_3)}$
from non-negativity of each increment, so the ordered levels cannot cross for
any input. The published estimate uses $\kappa^\star=\kappa_2$; the other
heads provide lower and upper operational levels trained with pinball losses. The axes cost $O(L_fWF^2d)+O(L_rW^2d)$ per
sample; with $F=54$, $W=20$, $L_f=2$, $L_r=4$ the feature axis exceeds the
sequence axis by about $73\times$ and dominates.

\section{Objective}
Internal customer outage survey analysis indicates that satisfaction responds
asymmetrically to error: underprediction, where restoration arrives later than
promised, reduces satisfaction sharply; overprediction is tolerated to a
threshold and penalised beyond. We encode that operational preference through
the following piecewise penalty template, with $\epsilon=\hat y-y$ in hours:
\begin{equation}
\ell(\epsilon)=
\begin{cases}
\alpha|\epsilon|, & \epsilon<0,\\
|\epsilon|, & 0\le\epsilon\le\theta,\\
\beta|\epsilon|, & \epsilon>\theta,
\end{cases}
\label{eq:loss}
\end{equation}
with $\theta=8$\,h, $\alpha=5$, and $\beta=2$ obtained from internal
customer outage survey analysis rather than tuned to model performance, and
held fixed for every method compared.

\begin{lemma}
Let $\kappa^\star=\alpha/(\alpha+1)$ and $\rho_\kappa(u)=u(\kappa-
\mathbb{I}[u<0])$ with $u=y-\hat y$. Restricted to $|\epsilon|\le\theta$,
\eqref{eq:loss} equals $(\alpha+1)$ times the pinball loss at $\kappa^\star$,
and $\kappa^\star$ is the unique level for which \eqref{eq:loss} and
$\rho_\kappa$ are proportional on that region.
\end{lemma}
\begin{proof}
Both losses are positively homogeneous and piecewise linear in $\epsilon$
with a single kink at $\epsilon=0$, so proportionality on
$|\epsilon|\le\theta$ holds iff the ratio of branch slopes agrees. For
$\epsilon<0$, $u=-\epsilon>0$, $\rho_\kappa(u)=\kappa|\epsilon|$ and
$\ell(\epsilon)=\alpha|\epsilon|$; for $0\le\epsilon\le\theta$, $u\le0$,
$\rho_\kappa(u)=(1-\kappa)|\epsilon|$ and $\ell(\epsilon)=|\epsilon|$. The
slope ratio of $\rho_\kappa$ is $\kappa/(1-\kappa)$ and that of $\ell$ is
$\alpha$; these agree exactly when $\kappa=\alpha/(\alpha+1)$, giving
uniqueness. Substituting $\kappa^\star$ yields
$\rho_{\kappa^\star}(u)=\alpha(\alpha+1)^{-1}|\epsilon|$ against
$\alpha|\epsilon|$ on the first branch and $(\alpha+1)^{-1}|\epsilon|$
against $|\epsilon|$ on the second; the constant is $\alpha+1$ on both.
\end{proof}

With $\alpha=5$, the slope ratio within the central operating region
corresponds exactly to $\kappa^\star=5/6$, motivating the primary level of
\eqref{eq:head2}. We retain the fractional representation so this local
correspondence holds without approximation. The resulting operating point
intentionally penalises underprediction more heavily than moderate
overprediction and therefore should not be interpreted as a
conditional-mean estimate.

\textbf{Smoothing and training.} Equation~\eqref{eq:loss} is an operational
penalty template with breakpoints at $\epsilon\in\{0,\theta\}$. Training
uses its Huber-smoothed surrogate $\ell_\delta$, which preserves the outer
linear branches and uses quadratic transitions within
$\delta$-neighbourhoods of the breakpoints, with $\delta=0.25$\,h. We write
$\rho_{\delta,\kappa}$ for the identically smoothed pinball loss. The same $\delta$ is used for every
method, including the boosting baselines, whose Newton updates need a
non-vanishing second derivative. A stability term
$\mathcal{L}^{\mathrm{stab}}$ penalises movement of the published timestamp
between consecutive revisions beyond $\tau_{\mathrm{tol}}=1$\,h, computed
under an event-grouped sampler. With
$\epsilon^{(\kappa)}_{i,j}=\hat y^{(\kappa)}_{i,j}-y_{i,j}$,
\begin{equation}
\mathcal{L}=\frac{1}{|S|}\!\!\sum_{(i,j)\in S}\!\!\Bigl[
\ell_\delta\bigl(\epsilon^{(\kappa^\star)}_{i,j}\bigr)
+\lambda_q\!\!\!\!\sum_{\kappa\in Q\setminus\{\kappa^\star\}}\!\!\!\!
\rho_{\delta,\kappa}\bigl(y_{i,j}-\hat y^{(\kappa)}_{i,j}\bigr)\Bigr]
+\lambda\mathcal{L}^{\mathrm{stab}}.
\label{eq:obj}
\end{equation}
Outside the smoothing neighbourhoods, $\ell_\delta$ retains the linear
branches of \eqref{eq:loss}; within them the quadratic transitions preserve
a nonzero penalty for systematic conservatism. The objective therefore has no
flat region.

\textbf{Scale.} The breakpoints of \eqref{eq:loss} are in hours; a monotone
transform such as $\log(1+y)$ would displace them to input-dependent
locations and destroy that interpretation, so we supervise on the hour scale.
The heavy tail is accommodated by linearity of $\ell$ outside the smoothing
neighbourhoods and by the constant $s$ of \eqref{eq:head2}, set to the
training-partition median remaining time per company so softplus arguments
are $O(1)$ at initialisation. Inputs are scaled by median and IQR on training
storms only. Configuration: $d=128$, $L_f=2$, $L_r=4$, $8$ heads, one context
cross-attention layer, $d_t=16$, dropout $0.1$, AdamW (decay $0.01$), lr
$3\times10^{-4}$ cosine with $500$-step warmup, batch $256$, $\lambda_q=0.2$,
$\lambda=0.1$, $1.3$\,M parameters.

\begin{table*}[t]
\centering
\caption{Five-seed test means by operating company, ordered by median revisions per outage and weighted by customer-hours
of exposure. W: weighted asymmetric error (h) $\downarrow$; R: RMSE (h)
$\downarrow$; C: satisfaction-rate statistic $\uparrow$ (read jointly with W
and R). Best per column in bold.}
\label{tab:main}
\scriptsize
\setlength{\tabcolsep}{2.6pt}
\begin{tabular}{l ccc ccc ccc ccc ccc ccc}
\toprule
& \multicolumn{3}{c}{OPCO-1} & \multicolumn{3}{c}{OPCO-2}
& \multicolumn{3}{c}{OPCO-3} & \multicolumn{3}{c}{OPCO-4}
& \multicolumn{3}{c}{OPCO-5} & \multicolumn{3}{c}{OPCO-6}\\
\cmidrule(lr){2-4}\cmidrule(lr){5-7}\cmidrule(lr){8-10}
\cmidrule(lr){11-13}\cmidrule(lr){14-16}\cmidrule(lr){17-19}
Method & W & R & C & W & R & C & W & R & C & W & R & C
& W & R & C & W & R & C\\
\midrule
Incumbent OMS & 56.57 & 23.33 & .610 & 19.49 & 7.82 & .554 & 7.93 & 5.35 & .817
 & 72.03 & 45.94 & .730 & 36.22 & 24.30 & .734 & 20.57 & 10.45 & .608\\
Climatology & 133.78 & 34.67 & .303 & 22.52 & 7.27 & .456 & 15.81 & 5.26 & .513
 & 87.02 & 22.47 & .293 & 49.96 & 13.92 & .349 & 41.77 & 11.46 & .381\\
LightGBM & 110.72 & 30.79 & .390 & 9.63 & 5.39 & .787 & 9.79 & 4.28 & .715
 & 64.43 & 18.57 & .429 & 29.12 & 10.08 & .578 & 16.28 & 7.64 & .715\\
XGBoost & 63.03 & 21.36 & .551 & 9.56 & 5.80 & .814 & 9.35 & 4.01 & .699
 & 40.67 & 14.78 & .610 & 24.56 & 9.35 & .634 & 18.00 & 8.26 & .703\\
CatBoost & 77.84 & 24.14 & .493 & 12.87 & 7.03 & .766 & 7.71 & 4.68 & .868
 & 54.57 & 16.79 & .519 & 22.49 & 9.33 & .703 & 21.03 & 10.91 & .710\\
LightGBM (lag) & 110.45 & 30.74 & .391 & 9.57 & 5.48 & .789 & 9.96 & 4.30 & .701
 & 63.86 & 18.47 & .435 & 29.00 & 10.07 & .576 & 16.53 & 7.67 & .694\\
GRU & 88.12 & 26.41 & .468 & 9.74 & \textbf{5.04} & .759 & 8.05 & 4.42 & .826
 & 49.69 & 15.75 & .549 & 25.24 & 10.02 & .654 & 19.55 & 9.44 & .708\\
LSTM & 88.37 & 26.62 & .472 & 9.84 & 5.58 & .784 & 8.26 & 4.64 & .825
 & 51.01 & 16.03 & .534 & 25.17 & 10.03 & .651 & 20.52 & 9.27 & .700\\
LTT (ours) & \textbf{50.91} & \textbf{19.15} & \textbf{.649}
 & \textbf{8.97} & 5.27 & \textbf{.830}
 & \textbf{6.59} & \textbf{3.46} & \textbf{.889}
 & \textbf{36.01} & \textbf{13.03} & \textbf{.757}
 & \textbf{19.99} & \textbf{8.27} & \textbf{.761}
 & \textbf{14.60} & \textbf{6.83} & \textbf{.757}\\
\bottomrule
\end{tabular}
\end{table*}

\section{Data}
We study six operating companies of a single holding company, OPCO-1 through
OPCO-6, dispatched independently under separate procedures and differing in
network construction, vegetation exposure and crew deployment. They are not
independent utilities and we treat them as correlated replicates. Results are
reported by anonymised company label under the data-sharing agreement.

\textbf{Cohort.} The event table intermixes populations with different
restoration dynamics, distinguished by an alarm-status prefix: momentary
interruptions resolve in seconds with no customer-facing estimate; planned
outages carry a scheduled restoration time; switching operations often
interrupt no customers; cancelled events record when a dispatcher closed a
duplicate report, not when supply returned. Retaining them would inflate
apparent accuracy for every method. We keep standard and SCADA-detected
customer outages, excluding cancelled and censored events, and apply the
filter identically to every method, so it affects the magnitude of errors but
not the ordering. Outages are attributed to storms by district and opening
timestamp, each window extended by five days for the restoration tail and
overlapping windows fused into one storm period. Splits and resampling are
defined at that granularity.

\textbf{Features and leakage control.} Each revision carries $47$ continuous
and $7$ categorical dynamic features plus $34$ contextual and $30$ static, in
three groups. \emph{Event state}: customers affected, critical customers,
priority, decoded alarm status. \emph{Lifecycle dynamics}: change in affected
customers, implied restoration rate, time since crew assignment, dispatch and
arrival, cumulative status regressions and crew suspensions. \emph{System
context}: concurrent open events, recent restoration throughput and
district-to-system strain, at company, district and feeder scope. All
features use information available at the revision timestamp; lifetime
quantities such as maximum customers affected are replaced by causal
counterparts accumulated to the current revision. Context aggregates use
strictly backward-looking windows and exclude the subject event from its own
aggregate at every scope, feeder included. We further verify leakage absence
by permuting restoration timestamps across outages within each storm.

\textbf{Characteristics.} Median restoration duration ranges from $3.45$ to
$6.33$\,h with ninetieth percentiles between $12.22$ and $39.15$\,h; at every
company the ninetieth percentile exceeds three times that company's own
median, and at one company it exceeds six times. The operational problem is
therefore to identify early the minority of outages that run long, not to
characterise the typical one. Revision counts differ substantially
(Table~\ref{tab:cohort}), reflecting both how often a company updates the
system and the granularity at which it opens events. Mean customers affected
per event varies by nearly an order of magnitude, so companies open events at
differing levels of network aggregation; exposure-weighted metrics are
therefore not strictly comparable across companies and we read differences
within one.

\textbf{Partitions.} Storms are assigned at storm granularity under a rule
fixed in a versioned configuration before any modelling: the holdout spans
the three magnitude classes and draws from distinct calendar months, so no
class or season is unrepresented. Random stratification was rejected because
the storm count per company is small enough that a random draw can omit a
magnitude class. The test partition holds nine storms at five companies and
three at OPCO-3, whose observation window is shorter. The assignment is not
chronological, so the setting is interpolation across storms rather than
forecasting of future ones.

\section{Evaluation and Setup}
\textbf{Point metrics.} An estimate seen by five thousand customers for six
hours is not equivalent to one corrected in ten minutes, so every metric is
weighted by customer exposure: revision $j$ of event $i$ carries
$w_{i,j}=c_id_{i,j}$, customers affected by event $i$ times the interval over
which that estimate remained published, the final revision running to
restoration. All are computed on $\hat y^{(\kappa^\star)}$. Writing
$\langle\cdot\rangle$ for the $w$-weighted mean, we report
$\mathrm{WAE}=\langle\ell(\epsilon)\rangle$,
$\mathrm{RMSE}=\langle\epsilon^2\rangle^{1/2}$, and
\begin{equation}
\mathrm{CSI}=1-\frac{\alpha\cdot\mathrm{UPR}+\beta\cdot\mathrm{OPR}_8}
{\alpha+\beta},
\label{eq:csi}
\end{equation}
with UPR the underprediction rate and $\mathrm{OPR}_8$ the overprediction
rate beyond $\theta$, both likewise $w$-weighted. The weighting concentrates
mass on the large, long-running events identified above, where an inaccurate
estimate is most costly.

CSI depends on two rates alone and is therefore insensitive to error
magnitude: underpredictions of two minutes and of twenty hours contribute
identically. We treat it as a satisfaction-rate statistic and never report it
alone; WAE is primary and RMSE the magnitude check. This also qualifies the
incumbent, since a systematically conservative estimator can attain a
favourable CSI without a favourable WAE or RMSE, which we test directly.
Because WAE and CSI share $\alpha,\beta$ with the training objective, every
baseline admitting a custom objective is trained on the identical loss, and
we report the ranking's sensitivity to $(\alpha,\theta)$.

\textbf{Baselines and controls.} We compare against the estimates the OMS
published during the same storms, a climatological predictor, boosted trees
in static and lag-flattened configurations, and recurrent models on identical
window tensors. All consume one shared feature construction, so only the
depth of history and the manner of its consumption differ, and every model
admitting a custom objective is trained on \eqref{eq:obj} under identical
scaling. Hyperparameters come from grouped cross-validation within training
storms, folds at storm granularity grouped by event, sixty trials per model,
and five seeds.

\textbf{Incumbent cadence.} The incumbent OMS and the learned models operate
at different update cadences. The OMS estimate is evaluated exactly as
published operationally: the most recent available ETR remains in effect
until it is replaced by a subsequent estimate. In contrast, all learned
baselines and LTT are permitted to issue an estimate at every OMS revision.
Exposure weighting therefore scores estimates according to the customer-hours
for which they remain in effect rather than according to the number of
predictions issued. The comparison with the incumbent measures end-to-end
operational performance, including the incumbent process cadence, whereas
comparisons among the learned baselines isolate predictive performance under a
common revision-level cadence. As an additional control, we score LTT using
the most recent model revision available at or before each OMS update point;
the median gain remains $34.2\,\%$.

\textbf{Lag-flattened baseline.} This gives a boosting model the four most
recent revisions concatenated with their first differences, testing whether
the benefit is obtainable by feature engineering. Since lag-$4$ against
$W\!=\!20$ confounds depth with architecture, we add a deeper lag-$K$ variant
and claim only that shallow lag features do not recover the benefit; to
separate an uninformative feature set from an unused one we report the share
of split gain accruing to lag features, $27.8\,\%$.

\section{Results}\label{sec:results}
\textbf{Main comparison.} LTT attains the lowest WAE at all six companies
(Table~\ref{tab:main}), reducing error against the incumbent by a median of
$36.9\,\%$ (range $10.0$--$54.0\,\%$) and against the strongest learned
baseline at each company by $11.3\,\%$ (range $6.2$--$19.2\,\%$).

\textbf{The incumbent's satisfaction score is bought with overprediction.}
The incumbent attains a competitive CSI at OPCO-1, OPCO-4 and OPCO-5, where
no learned method other than LTT matches it, yet at OPCO-4 and OPCO-5 its
RMSE exceeds that of every other method in Table~\ref{tab:main}, by factors
of $2.0$ and $1.7$ over the next worst. This pattern indicates systematic
conservatism, which CSI, as a rate statistic, cannot penalise. LTT is the
only method improving on the incumbent's CSI at all six companies while also
reducing RMSE at all six, and we regard this conjunction, not the CSI column
alone, as the operative result. For completeness, LTT does not attain the
best RMSE at OPCO-2, where GRU is lower ($5.04$ against $5.27$); it retains
the best WAE and CSI there.

\textbf{Operating-point coverage.} The primary level is intentionally
asymmetric rather than mean-seeking. On the pooled test predictions,
$87\,\%$ of realised remaining durations fall below the primary prediction,
indicating a mildly conservative operating point. We report this as a
descriptive coverage statistic rather than as a claim of exact conditional
quantile calibration.

\begin{figure}[t]
\centering
\includegraphics[width=.93\columnwidth]{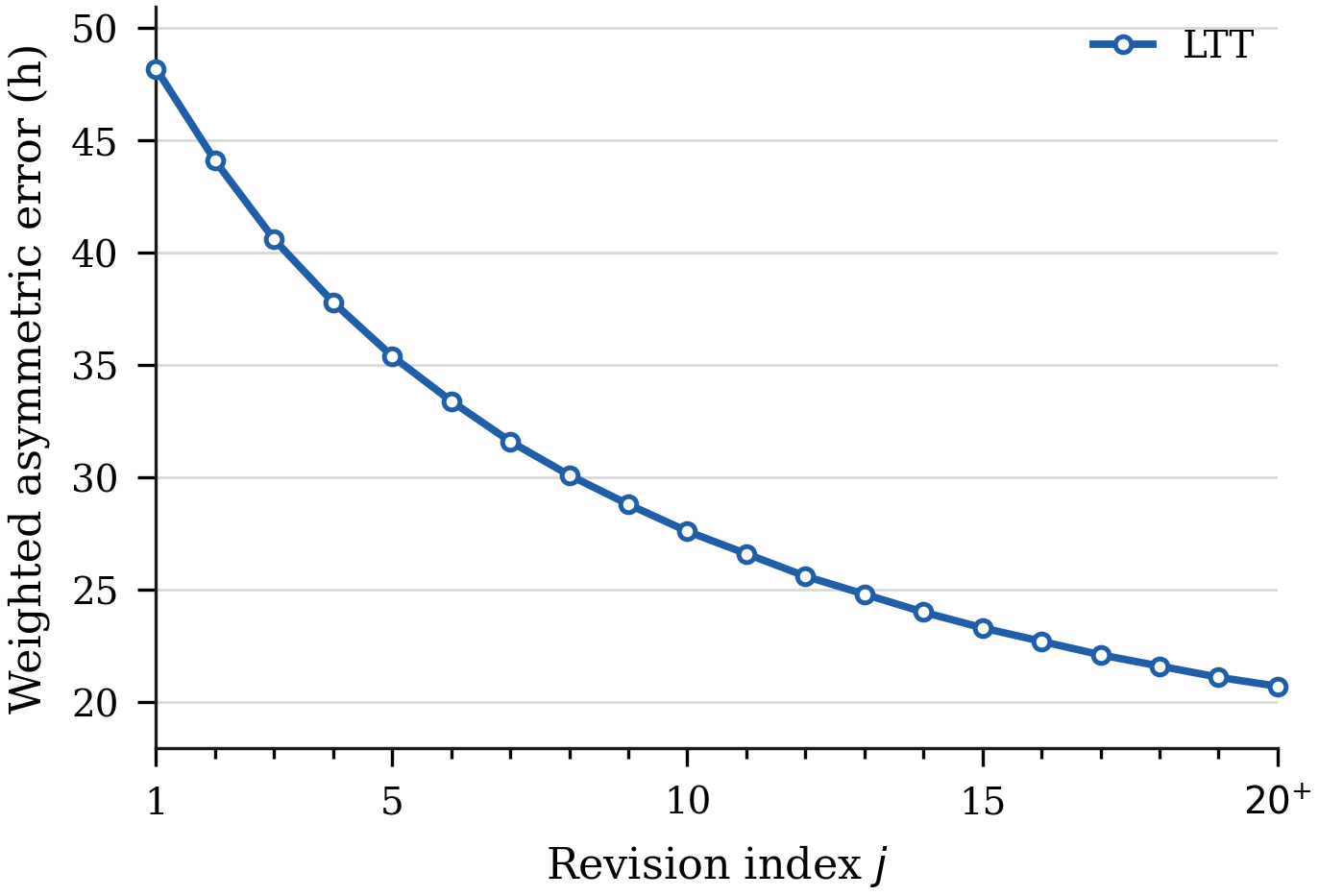}
\caption{LTT test error by revision index $j$, pooled over the six test
partitions and weighted by customer-hours of exposure as in Sec.~VI; the
final point aggregates $j\ge20$. Error is largest at the first revision,
where the window holds a single observed position and the model has no
history to condition on, and falls monotonically as revisions accumulate.}
\label{fig:revidx}
\end{figure}

\textbf{Error across the revision trajectory.}
Fig.~\ref{fig:revidx} stratifies LTT's error by revision index. It is largest
at $j=1$, where the window holds one observed position and there is no history
to condition on, and falls monotonically thereafter. This pattern is
consistent with the model benefiting from information accumulated along the
outage trajectory. The overall comparison in Table~\ref{tab:main} separately
shows that LTT achieves the lowest WAE among the learned methods at all six
operating companies.

\textbf{The gain is not explained by cadence alone.} LTT issues more estimates
than the OMS, so part of the end-to-end incumbent comparison can reflect the
operational value of fresher information. That does not explain the learned
baseline comparison, because all learned methods operate at the same
revision-level cadence and LTT still reduces WAE by a median of $11.3\,\%$
against the strongest learned baseline. Moreover, when LTT is scored only
using the most recent prediction available at or before each OMS update point,
its median gain over the incumbent remains $34.2\,\%$.

\textbf{Sensitivity and feature engineering.} Recomputing the ranking on
saved predictions over $\alpha\in\{3,5,8\}$, $\theta\in\{4,8,12\}$\,h leaves
the ordering unchanged, so the ranking is insensitive to the constants
derived from the internal customer outage survey analysis; models are not
retrained, so this is a sensitivity of the scoring rule. Lag-$4$ features change WAE by under
$1\,\%$ against plain LightGBM everywhere while receiving $27.8\,\%$ of split
gain, so they are used and uninformative rather than absent; a deeper lag-$K$
variant closes $31\,\%$ of the gap. We note that LightGBM and XGBoost differ
by $1.76\times$ in WAE at OPCO-1 despite identical features and budget, so no
single boosting implementation should be read as \emph{the} tabular
baseline.

\section{Conclusion}
Treating an outage as a sequence of revisions rather than a single record
reduces customer-weighted ETR error at all six operating companies studied,
by a median of $36.9\,\%$ against the estimates the utilities published
during the same storms and $11.3\,\%$ against the strongest learned baseline.
LTT error is largest at the first revision and decreases as information
accumulates across revisions. Its gain over the incumbent is not explained by
update cadence alone: all learned baselines share the revision-level cadence,
and a cadence-matched incumbent comparison retains a median $34.2\,\%$ gain.
The incumbent's favourable satisfaction rate reflects systematic
overprediction, which a rate statistic cannot penalise but RMSE exposes. Substantial error nonetheless remains at
every company. Characterising how much of it is irreducible, chronological
evaluation, and an aggregate restoration-curve measure are the subject of
ongoing work.

\balance

\end{document}